\documentclass[journal]{IEEEtran}
\usepackage{times}
\usepackage{epsfig}
\usepackage{graphicx}
\usepackage{amsmath}
\usepackage{amssymb}

\usepackage{cite}
\usepackage{enumitem}

\usepackage[ruled]{algorithm2e}

\usepackage{multirow}
\usepackage{makecell}
\usepackage{booktabs}
\usepackage{color}

\newcommand{\etal}{\emph{et al.}}
\newcommand{\ie}{\emph{i.e.}}

\ifCLASSINFOpdf
\else
\fi
\begin{document}
%
\title{Towards Class-agnostic Tracking Using Feature Decorrelation in Point Clouds}
%
%
%

\author{Shengjing~Tian,
        Jun Liu,
        and~Xiuping Liu
\thanks{Shengjing~Tian and Xiuping Liu are with the School of Mathematical Sciences, Dalian University of Technology, Dalian, 116024, China (e-mail: xpliu@dlut.edu.cn).}
\thanks{Jun Liu is with Information Systems Technology and Design Pillar, Singapore University of Technology and Design.}
\thanks{Manuscript received --; revised --}}

%
%

\markboth{Journal of \LaTeX\ Class Files,~Vol.~14, No.~8, July~2021}%
{Shell \MakeLowercase{\textit{et al.}}: Bare Demo of IEEEtran.cls for IEEE Journals}
%



\maketitle

\begin{abstract}
Single object tracking in point clouds has been attracting more and more attention owing to the presence of LiDAR sensors in 3D vision. 
However, existing methods based on deep neural networks mainly focus on training different models for different categories, which makes them unable to perform well in real-world applications when encountering classes unseen during the training phase.
In this work, we investigate     a more challenging task in the LiDAR point clouds, namely class-agnostic tracking, where a general model is supposed to be learned to handle targets of both observed and unseen categories.
In particular, we first investigate the class-agnostic performances of the state-of-the-art trackers via exposing the unseen categories to them during testing. 
It is found that when the distribution is shifted from observed to unseen classes, how to constrain the fused features between template and search region to maintain generalization is a key factor of class agnostic tracking.
Therefore, we propose a feature decorrelation method to address this problem, which eliminates the spurious correlations of the fused features through a set of learned weights, and further makes the search region consistent among foreground points and distinctive between foreground and background points. 
Experiments on the KITTI and NuScenes demonstrate that the proposed method can achieve considerable improvements by benchmarking against the advanced trackers P2B and BAT, especially when tracking unseen objects.
\end{abstract}

\begin{IEEEkeywords}
Point Clouds, Class-agnostic Tracking, Sample Weighting, Siamese Network.
\end{IEEEkeywords}

%
\IEEEpeerreviewmaketitle

\section{Introduction}
Single object tracking (SOT) in point clouds, also named 3D SOT, is a challenging yet meaningful research area, which has been sparked off increasing attention due to its potential applications in autonomous driving \cite{Argoverse2019, FaF2018}, robot vision \cite{Wang2017RobotsNavi}, and human-computer interaction \cite{annotation}. Given a specified template tightly enclosed by a 3D bounding box in the first frame, 3D SOT aims to estimate the state of the target (center, size, and yaw angle) relied on the geometric shape distribution. Recently, even through 2D visual tracking has made much progress with the benefits of deep learning \cite{MultiStream2020TIP}, it often cannot be directly lifted to the 3D scenes, on account of the sparsity and structure differences of point clouds compared to the 2D images. 

With the resurgence of deep learning, researchers take advantage of either convolutional neural network (CNN) or PointNet family \cite{PointNet2017,PointNetPP2017} to handle the 3D SOT task. The CNN-based methods compress the source point clouds into bird-eye-view (BEV) images to produce structured data, after which the standard 2D CNN can be implemented to predict the object state \cite{FaF2018}. Nevertheless, such BEV data can forfeit geometry information. In light of this, the recent mainstream \cite{Siam3D2019, P2B, BAT} utilizes PointNet \cite{PointNet2017} or PointNet++ \cite{PointNetPP2017} to extract features and devote to the embedding procedure between the template and search region within the Siamese network framework. The pioneer SC3D \cite{Siam3D2019} directly adopts the cosine similarity of the learned features to determine whether the candidate derived from Kalman filtering is the final target bounding box. Subsequently, P2B \cite{P2B} designs a feature fusion module by concatenating cosine similarity, coordinates and learned features, and BAT \cite{BAT} makes the network regress Box-Cloud representations to guide the next embedding procedure. 

\begin{figure}[t]
    \centering
    \includegraphics[width=1\linewidth]{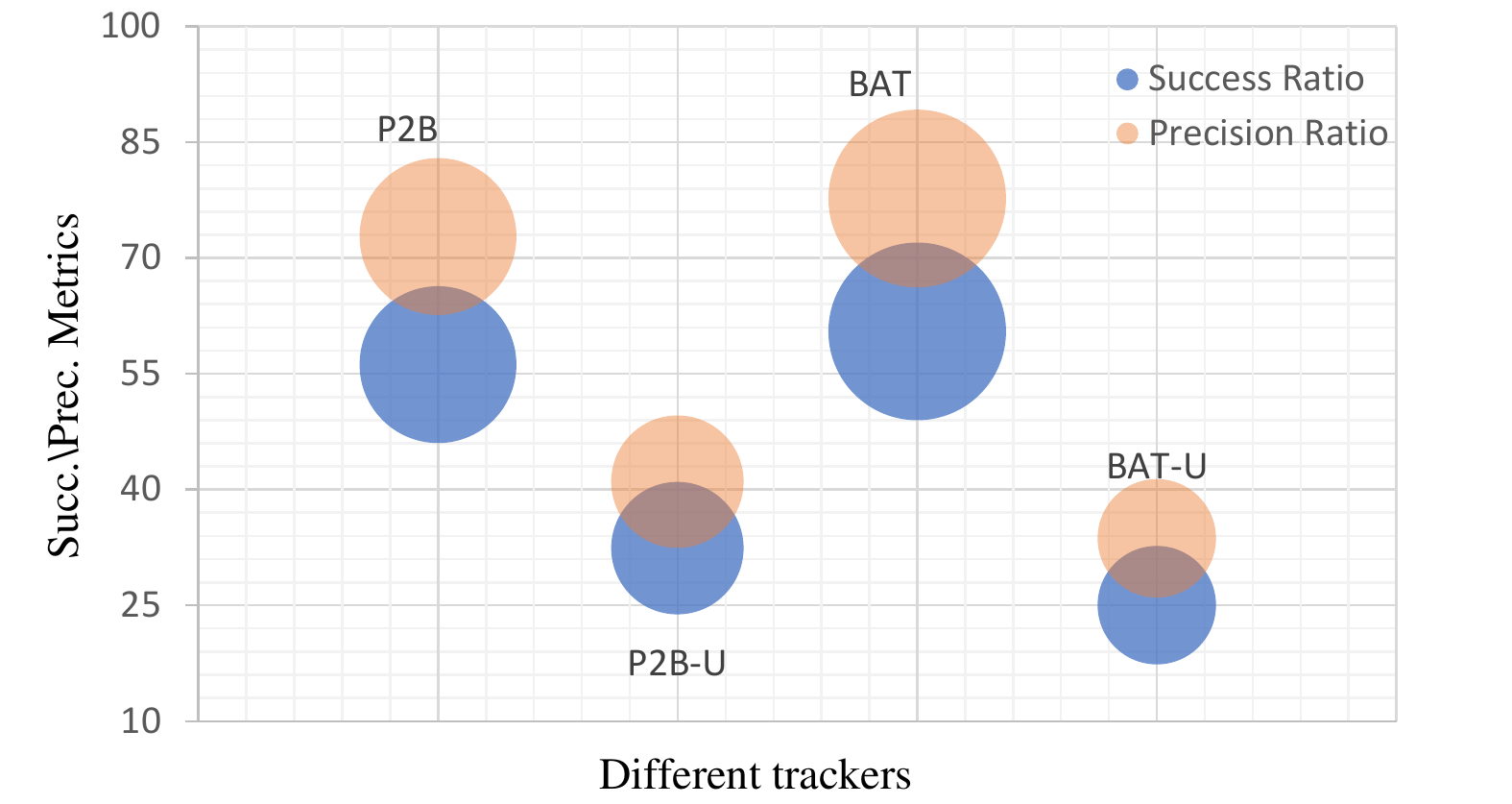}
    \caption{Comparisons between class-specific and class-agnostic tracking in 3D point clouds. We show the precision (orange) and success (blue) metrics. Here, P2B and BAT focus on the class-specific tracking which train and test only on cars. Differently, for the class-agnostic tacking, the methods with the suffix ``-U'' are also tested on cars though they do not observe this category during training phrase. There exists the significant gaps between P2B (BAT) and P2B-U (BAT-U).}
    \label{fig:intuition}
\end{figure}

Although achieving impressive performances, these PointNet-based methods are often inclined to class-specific tracking, which trains the model on a single class and evaluates it on the same category. In other words, they use the I.I.D. assumption. We argue that, in such a way, the Siamese network-based trackers are not able to exert its inherent merits and the samples of unobserved categories (or unseen target shapes) will reduce the tracking accuracy when applied to real-world scenarios. Generally, for single object tracking, the learned Siamese network is supposed to act as a general matching function that can track previously unseen targets once trained \cite{sint}. Therefore, class-agnostic tracking in point clouds is now an important problem to be solved.

\begin{figure}[t]
    \centering
    \includegraphics[width=1\linewidth]{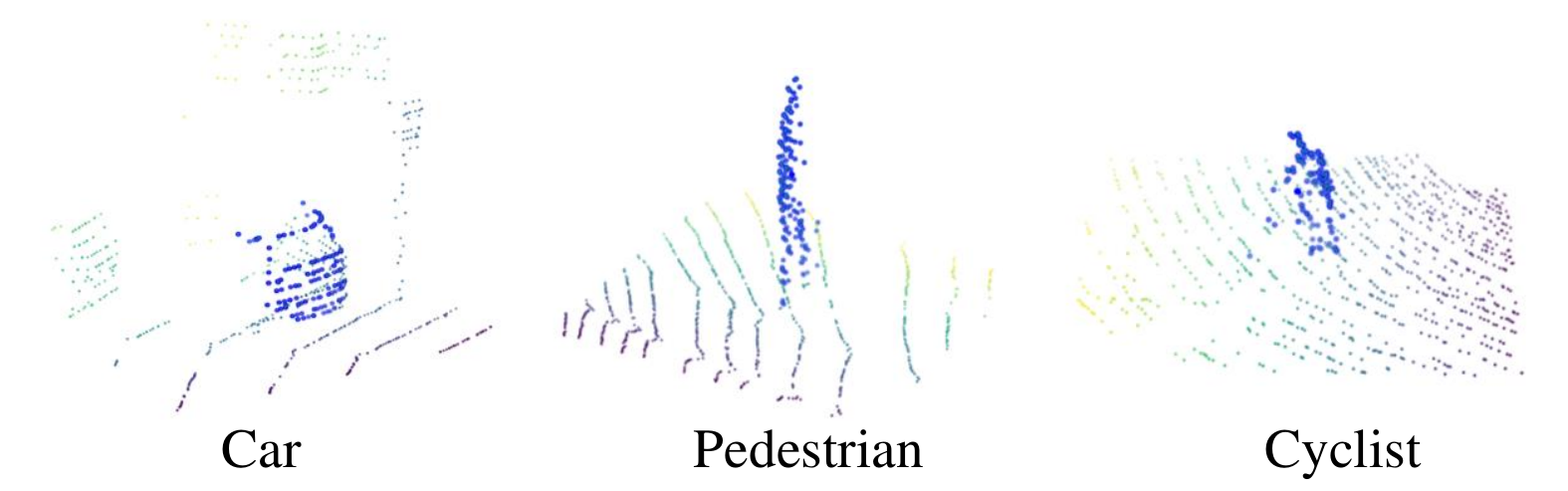}
    \caption{Different categories have large point distribution differences. In light of this, exposing unseen categories to trackers can help test their class-agnostic tracking ability.}
    \label{fig:category}
\end{figure}

Considering that recent 3D trackers generally lack specific designs for class-agnostic tracking, our focus has been shifted towards developing a generalization method for the 3D SOT. To grease the study of this problem, we first present new settings based on two large-scale LiDAR datasets: KITTI \cite{KITTI2013} and nuScenes \cite{nuscenes}. 
The standpoint of these settings is to expose unseen categories to trackers during the testing stage. 
It can help us to evaluate the class-agnostic tracking performance of the Siamese network-based trackers, because there are large spatial distribution differences among different categories (as shown in Figure \ref{fig:category}). In practice, with the help of these settings, we find that the bellwether of 3D trackers, P2B \cite{P2B} and BAT \cite{BAT}, significantly decrease in success and precision metrics (See Figure \ref{fig:intuition}). In light of this, it is very meaningful for class-agnostic tracking to explore the methods capable of restoring the general matching function property of the current trackers. 

To this end, we propose a feature decorrelation method to improve the class-agnostic performance for the recently advanced trackers.
Essentially, the recent methods including P2B and BAT are primarily composed of three main modules: feature extraction, target information embedding, and target proposal generation. Our motivation towards class-agnostic tracking is that no matter what category the target is, the fused features from the information embedding phase should represent the characteristic of foreground consistency and the points 
lying on the target's surface ought to be discriminative from those on the background, based on which the resulting features will facilitate the foreground points to cluster together in the proposal generation module.
However, the main barrier on the way is spurious correlation between the background point and the target's location when distribution is shifted from observed categories to unseen categories, and such correlation is mainly caused by the embedding process between the irrelevant features of the search region and relevant features of the template. 
Recent literature\cite{sw_nature} claims that stable learning from the causality perspective can be a breakthrough for more realistic problem settings. Therefore, to solve this problem, we consider each channel of the fused feature as a variable that controls either a foreground point or a background point. Because the dependence between variables majorly affect such correlation under distribution shifts, we use the feature decorrelation module to get rid of the dependence. It first transforms the fused features into Random Fourier Feature \cite{rrf} to circumvent the difficulty in measuring the dependence of non-linear deep features. Then, we minimize the Frobenius norm of the partial cross-covariance matrix to produce weights that is used for balancing samples. Finally, the whole procedure are implemented by alternatively and iteratively optimizing  the weights of samples and the parameters of network.


To sum up, our contributions are as follows:
\begin{itemize}[topsep=0.5pt, parsep=0.5pt]
  \item Considering the point cloud sequences for testing are not probably drawn from the same distribution as the training data, we are the first to concentrate on the distribution shifts in 3D SOT, aiming to restore its class-agnostic characteristic. To grease this study, we present new experiment setting.
  \item Dedicating to class-agnostic tracking, we propose the feature decorrelation method for removing spurious correlations between the irrelevant features of the search region and relevant features of the template.
  \item On the large-scale datasets KITTI and NuScenes, the state-of-the-art methods, P2B and BAT, achieve better generalization ability employing the designed sample weighting, compared with their vanilla version.   
\end{itemize}

\section{Related Work}
\subsection{Single Object Tracking in Point Cloud.}
LiDAR point cloud characterizes by long-range spatial information, which has promoted a deep research history in 3D tracking. 
Held \etal \cite{Held2013PrecisionTW} constructed an up-sampled point cloud with color information, and then used a color-augmented alignment strategy to estimate the velocity of the tracked vehicle.  
Dewan \etal \cite{MotionBased} utilized the RANSAC algorithm to estimate motions of sensor and moving objects, and tracked objects by updating motion based on the Bayesian approach. However, these methods resort to data association or segmentation and just obtain object points instead of 3D bounding box indicating size and orientation. 
Subsequently, Asvadi \etal \cite{rgb_lidar2016} proposed a parallel mean-shift algorithms to predict 3D bounding box representation, and Li \etal  \cite{SOTracker} introduced an optimization-based algorithm that takes shape and motion priors into account. 

Recently deep learning brings new energy for 3D object tracking in point clouds, and some data-driven methods have been achieving impressive performance. For example, as a counterpart of 2D visual tracking using Siamese network \cite{SiamFC2016}, SC3D \cite{Siam3D2019} is the pioneer to learn a discriminative feature space for calculating cosine similarity between template and candidates. However, this method has the following drawbacks: 1) only using global feature embedding ignores the object relations in the search region; 2) the candidate generation is limited by the probability sampling; 3) the state estimation is not considered. Accordingly, Feng \etal \cite{retracker2020} and Tian \etal \cite{LearningIW2021} introduced coarse-to-fine and warp learning strategies to adjust the state of the tracked vehicle, respectively. 
Afterward, a milestone work in this area is that P2B \cite{P2B} takes inspiration from the deep Hough voting \cite{DeepHough2019}, and designs a feature embedding module to fulfill the Siamese-based region proposal network (RPN) \cite{SiamRPN2018} in 3D case. Taking the P2B as the baseline, Tian \etal \cite{DSDM} integrated the philosophy of supervised descent method into PointNet++ to refine the predicted bounding boxes progressively, Hui \etal \cite{V2B} substituted the point-based RPN head with a BEV-based module. Cui \etal \cite{LTTR} and Shan \etal \cite{PTT2021} leveraged the popular transformer to promote information embedding procedure by attention. Considering that mostly single-view scanned data have similar patterns but different sizes, BAT \cite{BAT} exploits the Box-Cloud representation as an additional clue to guide feature embedding. All these methods focus only on class-specific tracking, whereas we aim to put forward a class-agnostic tracking approach in point clouds.

\subsection{Siamese network on Point Cloud}
Some point cloud tasks based on the deep Siamese network have obtained impressive performances. In particular, point cloud registration is the most related topic with 3D SOT, which seeks the best transformation matrix to align the source and target point clouds.
Recently, Aoki \etal \cite{pointnetLK2019} incorporated the Lucas \& Kanade algorithm with PointNet \cite{PointNet2017} to devise a recurrent trainable network, called PointNetLK. Wang \etal \cite{dcp} used the PointNet to generate soft corresponding points and then solve the rigid transformation by a differentiable singular value decomposition. Choy \etal \cite{dgr} used a sparse convolution network to propose deep global registration with correspondence confidence prediction. Huang \etal \cite{predator} presented an overlap-attention block based on a graph neural network, dedicated to the registration problem with low overlap. As for the partially scanned objects in the wild, Gro{\ss} \etal \cite{alignnet} proposed AlignNet to learn to align scans in a coarse-to-fine manner. 
Actually, such Siamese network-based approaches also involve the generalization ability on unseen categories but are overlooked by them. Thus, based on the PointNetLK, Li \etal \cite{pointnetLK_revisit} claimed that the Jacobian matrix with respect to the input coordinates can improve the generalization ability for point cloud registration. Although the fusion and matching technologies that appear in these Siamese networks can provide some useful references to 3D SOT, how to implement a general tracking algorithm is still an open problem to be solved.

\subsection{Feature Decorrelation}
Towards the real-world applications, some works have paid attention to eliminating the spurious association between variables during the training. For example, with the decorrelation regularizer, Kuang \etal \cite{sw_kuang} constrained the generation of learned weights to improve the stability of the linear regression model. Shen \etal \cite{sw_shen} theoretically explained the problem occurring in distribution shifts between training and testing sets. Afterward, Shen \etal \cite{sw_dvd} proposed group-based variables partition to alleviate the problem of the over-reduced sample size. Nevertheless, these works are aimed at the linear model.  
He \etal \cite{sw_nico} thus proposed a Non-I.I.D. dataset for images recognition, and presented a batch balancing method. 
Furthermore, Zhang \etal \cite{sw_image} leveraged Random Fourier Features and iterative weight update to address the distribution shift in images classification. 
Although these methods have shown the effectiveness by disentangling the features, there are still no attempts that handle the distribution shifts for 3D tracking in point clouds. 
In our work, we present a feature decorrelation module for class-agnostic tracking, which is specially designed for the information embedding state. 
\section{Methodology}
In this work, we aim to handle class-agnostic tracking in point cloud. Below we first formulate the 3D SOT problem mathematically. Then, by taking the state-of-the-art trackers as baselines, we introduce a feature decorrelation method to improve the ability of tracking unseen targets. Finally, we describe new settings which are created upon two widely-used LiDAR datasets.

\begin{figure}[t]
    \centering
    \includegraphics[width=1\linewidth]{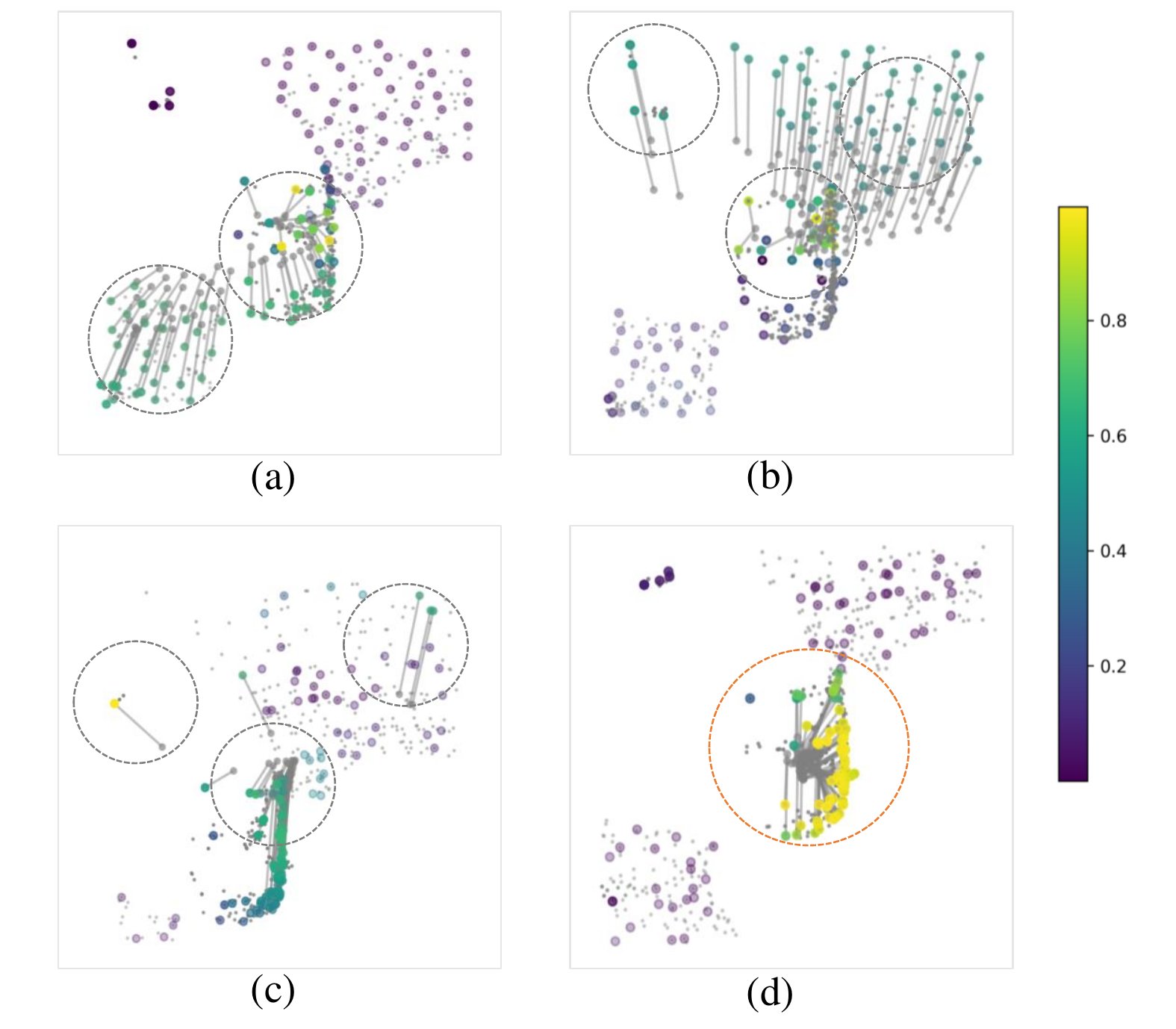}
    \caption{The effect of the fused features $Z$ generated by the information embedding module. We use different colors to indicate the possibility that each point belongs to the tracking target. The lower the probability, the darker the color. The gray lines represent voting results, which reflect the contribution of each point for the final tracked box. (a)-(c) show three cases of spurious correlation, which lead to some undesirable clusters (highlighted by gray circle) and spread-out voting results. (d) is an ideal case where there remains consistency among the foreground points and discrimination between the background (dark) and foreground (yellow) points.}
    \label{fig:spurious_correlation}
\end{figure}

\begin{figure*}[t]
    \centering
    \includegraphics[width=1\linewidth]{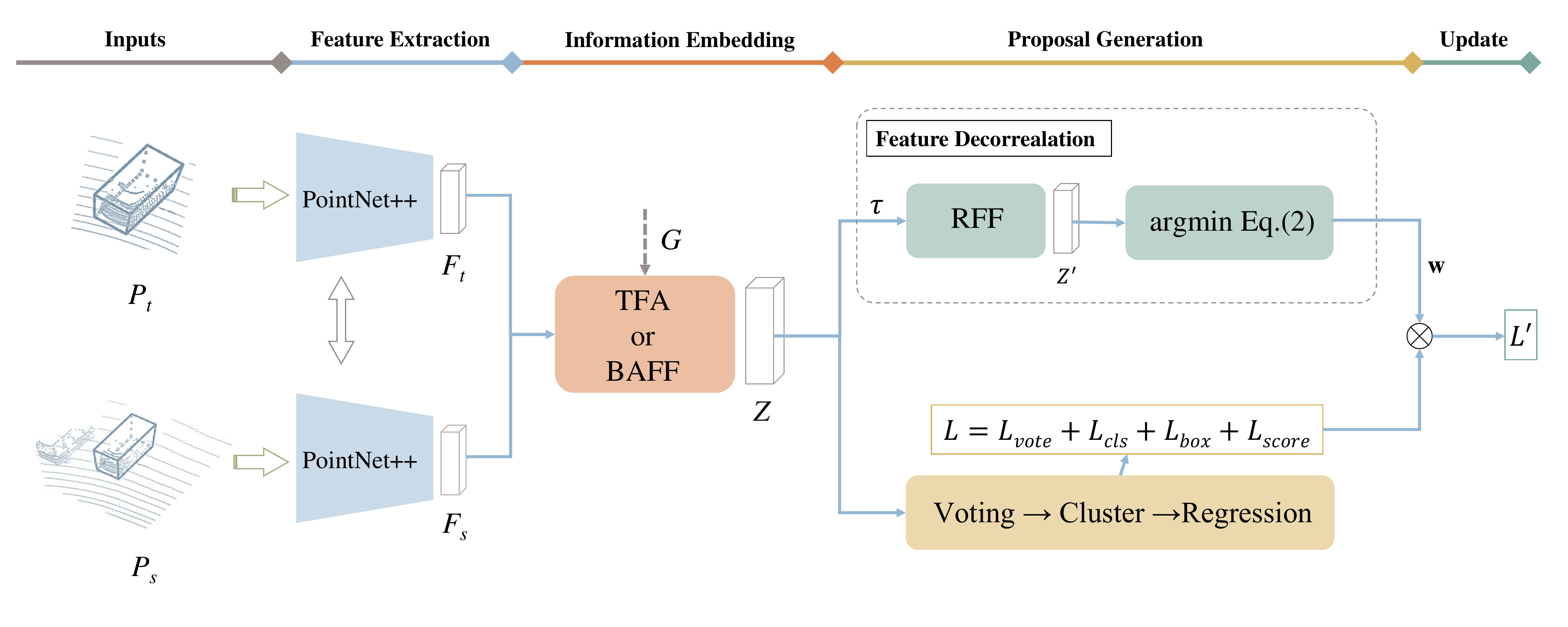}
    \caption{Pipeline of the proposed method. The current 3D trackers often contain the modules of feature extraction, target information embedding, and proposal generation. The inputs of $P_T$ and $P_S$ are first sent to PointNet++ to extract features. Then, the baseline generates the fused  features between the template and search region. Here, P2B uses the target-specific feature augmentation module (TFA), and BAT adopts the Box-aware feature fusion (BAFF). To make the fused features class-agnostic, we propose a feature decorrelation module based on the sample weighting method. Through alternative and iterative optimization, the resulting weights are then applied to the loss function computed in the proposal generation to make the points on the target surface more consistent. }
    \label{fig:framework}
\end{figure*}
\subsection{Problem Definition}
Let $P=\{p_i=(x_i, y_i, z_i)\}_{i=1}^N$ represent a point cloud with $N$ points captured by LiDAR. Given a specified target $P_T$ (also dubbed as template) in the first frame, 3D SOT aims to generate a 3D bounding box to accurately localize the target in each coming frame. In practice, considering that each scan frame covers a very large scenario, we usually crop a search region $P_S$ around the previous tracking result. Here, the template $P_T$ and search region $P_S$ are composed of $N_T$ and $N_S$ points, respectively. In addition, 3D bounding box $\mathcal{B}$ is used to represent the state of the target because it includes the center position $(x, y, z)$, size $(l, w, h)$ and yaw angle $\theta$, \ie, $\mathcal{B}=(x, y, z, l, w, h, \theta)$. Briefly Speaking, during online tracking, the proposed method takes the template $P_T$ and the search region  $P_S$ as input and predicts a bounding box $\mathcal{B}$ for the target in each frame.

\subsection{Baseline}
As a milestone, P2B \cite{P2B} proposes a point-to-box network for class-specific tracking, attracting much attention in the research forum of 3D SOT. Based on the P2B, there are many improvements \cite{DSDM, BAT, PTT2021, V2B, 3D-SiamRPN}, among which BAT \cite{BAT} introduces geometric information, Box-Cloud, to achieve considerable results. Therefore, we take these two representative trackers (P2B and BAT) as the baselines. 

Mathematically, let $\psi$ be the feature extractor, which can be PointNet++ \cite{PointNetPP2017}. The intermediate features of the template $P_T$ and search region $P_S$ can then be denoted as $F_T=\psi(P_T) \in R^{M_T \times D}$ and $F_S=\psi(P_S) \in R^{M_S \times D}$, respectively. Here, $M_T$ and $M_S$ are the number of the down-sampled points. The information embedding module $\phi$ consumes $F_T$, $F_S$ and other geometric information $G$, and then produces fused features $Z = \phi(F_T, F_S, G)$. As for P2B,  $G$ is point coordinates, \ie, $G=(P_T, P_S)$. For BAT, $G$ consists of template coordinates $P_T$ and its Box-Cloud $C_T \in R^{N_T \times 9}$, \ie, $G=(P_T, C_T)$. Finally, the proposal generation module $\gamma$ utilizes the deep Hough voting mechanism \cite{DeepHough2019} to output candidate bounding boxes $\{\hat{\mathcal{B}}_i\}_{i=1}^{N_{\mathcal{B}}}$ and its corresponding confidence scores $\{s_i\}_{i=1}^{N_{\mathcal{B}}}$. The target location is determined by $\hat{\mathcal{B}}_i$ with the highest score. 

To train the network, the loss function has the four constraints, \ie, 
\begin{equation}
    \mathcal{L} = \mathcal{L}_{vote} + \mathcal{L}_{cls} + \mathcal{L}_{box} + \mathcal{L}_{score}. 
\end{equation}
Concretely, $\mathcal{L}_{vote} = \frac{1}{M_S} \sum_{i=1}^{M_S} \mathbb{I}(p_i) \| p^* - (p_i + \Delta p_i) \|$ is responsible for deep Hough voting, where $p^*$ is the ground-truth of target center, $\Delta p_i$ is the predicted offset, and $\mathbb{I}(p_i)$ is the indicator function that outputs 1 if $p_i$ belongs to the target point, otherwise 0 to non-target point. 
$L_{cls}=-\frac{1}{M_S} \sum_i \mathbb{I}(p_i) \log(r_i)$ is for foreground classification, where $r_i$ is generated by an MLP based on $Z$. 
In addition, the predicted boxes and confidence score are supervised by $L_1$ smooth function and binary cross entropy, respectively. That is, 
$L_{box} = \frac{1}{N_{\mathcal{B}}} \sum_{i} \|\hat{\mathcal{B}}_i - \mathcal{B}^* \|$, where $\mathcal{B}^*$ is ground-truth;
$L_{score} = -\frac{1}{N_{\mathcal{B}}} \sum_{i} \mathbb{I}(\hat{\mathcal{B}}_i, \mathcal{B}^*) \log(s_i)$, where $ \mathbb{I}(\hat{\mathcal{B}}_i, \mathcal{B}^*)$ is 0 when the interaction-of-union between $\hat{\mathcal{B}}_i$ and $\mathcal{B}^*$ is less than 0.3, and 1 when it is larger than 0.6.

\subsection{Feature Decorrelation Method}\label{sec:sw}
The 3D SOT task can often be divided into three modules: feature extraction, target information embedding, and proposal generation.
Currently, most class-specific trackers focus on how to conduct the fusion between the template and search region in observed distribution, but ignore the generalization when confronted with out-of-distribution (OOD) tracklet. 
The main gap is that when tracking the OOD instances during testing stage, the background points usually have spurious correlations with the target’s location (see Fig.\ref{fig:spurious_correlation}-(a)(b)(c)).
This correlations mainly comes from the fused features trained on limited classes, which easily embed template's features into irrelevant features of search region of unseen category.

Therefore, our original motivation is to keep the two basis properties for class-agnostic tracking, regardless of whether the tracked class was observed or not. 
One of the properties is consistency which requires that the foreground features of the target should be as close as possible. The other one is discrimination which requires that the features between the foreground and background need to be as far away as possible. When the fused features belong to such pattern (see Fig.\ref{fig:spurious_correlation}-(d)), the last module, \ie, proposal generation, can learn to assemble the foreground points together and generate a good bounding box. 
Intrinsically, to attain this goal, the core lies in how to alleviate the correlation. Recently, the literature \cite{sw_nature, sw_image} points out this spurious correlations are caused by the statistical dependence among features. Along this mentality, we try to eliminate the dependence in fused features and propose a feature decorrelation method for class-agnostic tracking, which treats each channel of the fused features as a variable and removes the variable dependence via a set of learned sample weights.

\begin{figure}[t]
    \centering
    \includegraphics[width=1\linewidth]{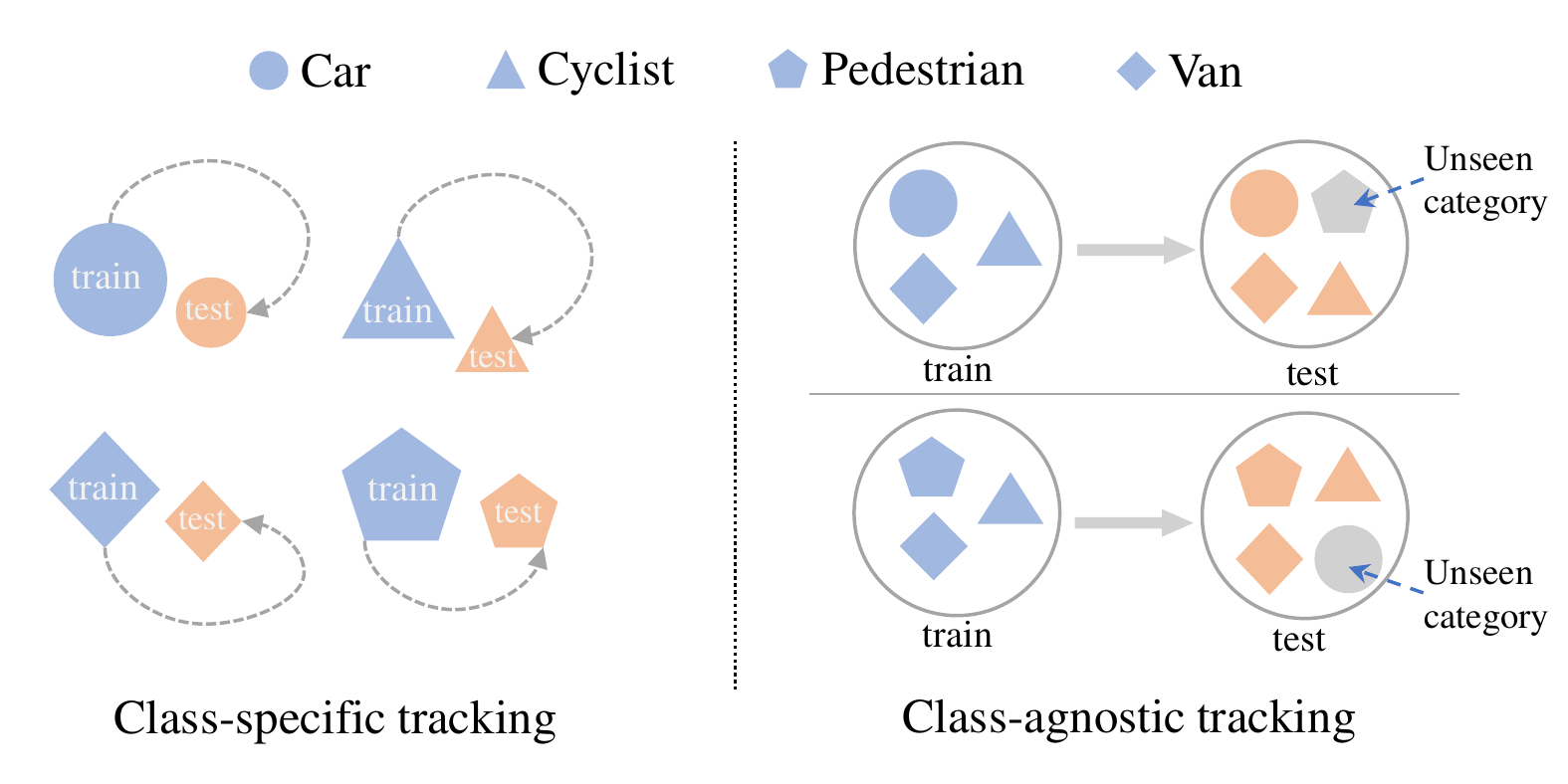}
    \caption{Illustration of class-specific and class-agnostic tracking. The circles, triangles, pentagons, and diamonds represent cars, cyclists, pedestrians, and vans, respectively. The training and testing sets are marked in blue and orange, respectively. The gray shape indicates it is not observed during training phase. As shown, the class-specific tracking trains and tests the models on the same category, whereas the class-agnostic tracking requires to evaluate the models on both observed and unseen categories.}
    \label{fig:setting}
\end{figure}
\begin{algorithm}[t]
\caption{Offline Training Pipline.}
\label{algo:offline_training}
\LinesNumbered
\KwIn{Template:$P_T$, Search Region: $P_S$,  and $\mathcal{B}$ for $P_T$}
\KwOut{Trained Network: $ \phi, \psi, \zeta$}
Initialize $ \phi^{(0)}, \psi^{(0)}, \zeta^{(0)}$\\
\For{iteration number $t$}{
    Get features: $F_T\gets\psi^{(t-1)}(P_T)$, $F_S\gets\psi^{(t-1)}(P_S)$\\
    Embed information:  $Z \gets \phi^{(t-1)}(F_T, F_S, G)$\\
    Estimate box and score: $(\mathcal{B}, s) \gets \gamma^{(t-1)}(Z)$\\
    // Decorrelate feature \\
    $Z^{\prime} = \tau(Z)$; \\
    \For{$i=1, \cdots, D$}{
        \For{$j=1, \cdots, D$}{
            \If{$j>i$}{
                calcullate $\hat{\Lambda}_{Z^{\prime, (t-1)}_i, Z^{\prime, (t-1)}_j}$
            }
        }
    }
    $\mathbf{w}^{(t-1)} \gets $ optimizing Eq.(\ref{eq:iter_w});\\
    $\phi^{(t)}, \psi^{(t)}, \zeta^{(t)} \gets$ optimizing Eq. (\ref{eq:iter_net})
}
\end{algorithm}
In detail, we first transform the fused features into $Z^{\prime} = \tau(Z)$, where $\tau: R^{M_S \times D} \rightarrow R^{D}$ is the aggregation function on the search region. In this work, we simply utilize \textit{Average Pooling} as $\tau$. Next, we can use the transformed features $Z^{\prime}$ to calculate sample weights. 
Unlike the linear model used in the previous methods \cite{sw_kuang, sw_dvd, sw_shen}, we deal with the non-linear features extracted by deep learning whose dependence is challenging to be measured. In particular, we adopt Random Fourier Features (RFF) \cite{rrf} for each variable $Z^{\prime}_{i}$, inspired by \cite{sw_image}. Theoretically, to test the independence of two variables $Z^{\prime}_{i}$ and $Z^{\prime}_{j}$, we need to judge whether the cross-covariance operator $\Lambda_{Z^{\prime}_{i}, Z^{\prime}_{j}}$ with respect to $u(Z^{\prime}_i)$ and $v(Z^{\prime}_j)$ is 0, where $u$ and $v$ are the element of Reproducing Kernel Hilbert Space $\mathcal{H}$. This is gurateened by the following proposition \cite{sw_image}: \textit{Assuming $\kappa_{Z^{\prime}_i}$ and $\kappa_{Z^{\prime}_j}$ are measurable, positive definite kernels defined on the variable $Z^{\prime}_i$ and $Z^{\prime}_j$ respectively, if the product $\kappa_{Z^{\prime}_i}\kappa_{Z^{\prime}_j}$ is characteristic, $\mathbb{E}(\kappa_{Z^{\prime}_i}(\cdot,\cdot)) < \infty$, $\mathbb{E}(\kappa_{Z^{\prime}_{j}}(\cdot,\cdot)) < \infty$, we have $\Lambda_{Z^{\prime}_i, Z^{\prime}_j}=0 \Leftrightarrow Z^{\prime}_i \perp Z^{\prime}_j$.}

RFF can provide such function space (denoted as $\mathcal{H}_{RFF}$) to  attain this goal. More specifically, in a mini-batch, we can have $B$ samples $(Z^{\prime}_{1,i}, Z^{\prime}_{2,i}, \dotsc, Z^{\prime}_{B,i})$, which are derived from the distribution of variable $Z^{\prime}_{i}$. Let $\mathbf{w}=(\omega_1, \dotsc, \omega_B)$ represent the sample weights to be learned, and satisfy $\sum\omega_m=1$. The weighted cross-covariance matrix can be estimated as follows:

\begin{small}
    \begin{equation}
        \begin{aligned}
            \hat{\Lambda}_{Z^{\prime}_{i}, Z^{\prime}_{j}} = \frac{1}{B-1} \sum\limits_{m=1}^{B} \Bigg[ &\left(\omega_m \mathbf{u}(Z^{\prime}_{m,i}) - \sum_k \omega_k \mathbf{u}(Z^{\prime}_{k,i}) \right) ^{\top} \\
            &\left(\omega_m \mathbf{v}(Z^{\prime}_{m,j}) - \sum_k \omega_k \mathbf{v}(Z^{\prime}_{k,j}) \right)\Bigg],
        \end{aligned}
    \end{equation}
\end{small}
where $\mathbf{u}=(u_1, u_2, \dotsc, u_{N_u})$ with $u_i \in \mathcal{H}_{RFF}, i=1, \dotsc, N_u$ and $\mathbf{v}=(v_1, v_2, \dotsc, u_{N_v})$ with $v_i \in \mathcal{H}_{RFF}, i=1, \dotsc, N_v$. Thereafter, we can optimize the Frobenius norm of the $\hat{\Lambda}_{Z^{\prime}_{i}, Z^{\prime}_{j}}$ to solve $\omega_m$,

    \begin{equation}\label{eq:w}
        \mathbf{w^*} = \arg \min \limits_{ \mathbf{w}} \sum\limits_{i=1}^D \sum\limits_{j > i}^D \left\| \hat{\Lambda}_{Z^{\prime}_{i}, Z^{\prime}_{j}} \right\|_F^2.
\end{equation}

The learned weights $\mathbf{w^*}$ are then applied into the final loss function to balance the distribution of each variable in the fused feature, which will enable the fused feature to be foreground consistent. Finally, the whole procedure can be summarized as alternative and iterative optimization (See Algorithm \ref{algo:offline_training}):
\begin{align}
    \mathbf{w}^{(t-1)} =  \arg \min \sum\limits_{i=1}^D \sum\limits_{j > i}^D \left\| \hat{\Lambda}_{Z^{\prime, (t-1)}_{i}, Z^{\prime, (t-1)}_{j}} \right\|_F^2; \label{eq:iter_w}\\
    \phi^{(t)}, \psi^{(t)}, \zeta^{(t)} = \arg\min \sum\limits_{m=1}^B \omega^{(t-1)}_m \mathcal{L}, \label{eq:iter_net}
\end{align}
where $Z^{\prime, (t-1)}$ is the fused features calculated by the network parameters of the last epoch, \ie,
\begin{equation}
    Z^{\prime, (t-1)} = \phi^{(t-1)} \left(\psi^{(t-1)}(P_T), \psi^{(t-1)}(P_S), G \right).
\end{equation} 

In this way, we can reduce the dependence on background points in the information embedding process, so that the fused features remain consistency and discrimination, and then perform class-agnostic tracking.

\subsection{Settings for Class-agnostic Tracking in Point Clouds} \label{sec:setting}
To promote the development of single object tracking in point clouds, Giancola \etal \cite{Siam3D2019} created a benchmark: KITTI Tracking dataset. On each category, they selected some sequences as training set and others as testing set. Following this, there have been many efforts \cite{retracker2020, P2B, DSDM, BAT, PTT2021, V2B} to improve the success and precision ratios on the dataset. It is worth noting that all these methods lack the specific designs for class-agnostic tracking and mainly focus on training different models for each category. However, as an extension of 2D visual tracking, a 3D SOT method should also satisfy the characteristic that it is capable of tracking any object as long as the target is specified in the first frame. For understanding, we illustrate class-specific and class-agnostic tracking in the Figure \ref{fig:setting}.

For class-agnostic tracking, we introduce four settings according to the mainstream datasets for LiDAR point clouds. Specifically, on KITTI, we use samples from four categories, including cars, pedestrians, vans and cyclists. On the first setting, we use the samples of pedestrians, vans and cyclists to train the models, and during the testing stage, the samples of cars are exposed to the models in addition to the three observed categories. For a well-round comparison, we swap the roles of cars and pedestrians in the second setting. 
Analogously, on NuScenes, we construct another two settings based on the categories of pedestrians, cars, trunks and bicycles. The first evaluates the class-agnostic performance by unseen category of cars while the second by pedestrians. We may want to traverse all permutations of different categories to obtain a more comprehensive experimental setup. In fact, cars and pedestrians account for the vast majority in the datasets, and these two classes have obvious distribution discrepancies. Therefore, it has been representative to take them as the unseen domain to measure the class-agnostic ability of 3D trackers. We also summarize those class-agnostic settings in the Table \ref{tab:setting}. 
 \renewcommand{\arraystretch}{1.5} 
\begin{table}[t]
  \setlength{\abovecaptionskip}{0.1cm}
  \setlength{\belowcaptionskip}{-0.cm}
  \caption{Experimental settings on KITTI and NuScenes. The context in the parentheses is the number of samples in its testing set.}
  \label{tab:setting}
  \centering
  \begin{tabular}{c|cccc}
    \toprule
        \multirow{2}{*}{ \quad} & \multicolumn{2}{c}{KITTI} & \multicolumn{2}{c}{NuScenes}\\
        \cmidrule(lr){2-3} \cmidrule(lr){4-5}
          &Setting-1 &Setting-2 &Setting-1 &Setting-2 \\
    \midrule
        \makecell{Observed \\ Category} & \makecell{Pedestrian\\ Van\\ Cyclist\\ (7644)} & \makecell{Car\\ Van\\ Cyclist \\(7980)} &\makecell{Pedestrian\\ Truck\\ Bicycle\\ (49106)} & \makecell{Car\\ Truck\\ Bicycle\\(80038)} \\
    \midrule
        \makecell{Unseen\\ Category} & \makecell{Car\\ (6424)}& \makecell{Pedestrian\\(6088)} & \makecell{Car\\(64159)} & \makecell{Pedestrian\\(33227)} \\
    \bottomrule
  \end{tabular}
\end{table}

\section{Experiments}
In this section, we first introduce the datasets and metrics that we used for 3D tracking. Next, the implementation details are described. Then the preliminary experiments are conducted to illustrate the shortcoming of the recently advanced trackers, Finally, we report the main comparison result and give some key analysis and visualization. 

\subsection{Datasets and Evaluation Metrics}
\textbf{Datasets.}
Following previous methods, we adopt the most popular dataset for LIDAR point cloud, KITTI Tracking Set\cite{KITTI2013}. Its samples are captured under 21 different scenes, totally yielding 41,713 frames and 808 tracking instances for training and testing. The target categories in this dataset mainly includes cars, pedestrians, vans, and cyclists. The coordinates of each sample are stored in terms of the camera coordinate system. 

In order to avoid over-fitting on a single dataset, we also consider another large-scale dataset, NuScenes\cite{nuscenes}, which is composed of 1,000 scenes. The coordinates of each sample are stored in terms of the LiDAR coordinate system. This dataset is very challenging because it gathers abundant samples from different traffic situations, weather conditions, vehicle types, and so on. Considering that the ground-truth of the testing data is not accessable, we evaluate the trackers on its validation set.

\textbf{Evaluation Metrics.}
Like 2D visual tracking, the performance metrics for 3D tracking are also success and precision ratios \cite{OTB2015}. The difference with 2D tracking lies in the calculations are extended from 2D to 3D. In particular, the success ratio is related to the interaction-of-union(IoU) between the predicted bounding box and ground-truth bounding box. It statistics the percentage of frames where the IoU is larger than the given thresholds. The precision ratio is dependent on the center distance (CD) between the predicted bounding box and ground-truth bounding box. It also calculates the percentage of frames where the CD is smaller than the given thresholds. By densely drawing the thresholds on the X-axis, we can obtain two specific curves whose Y-axis are success ratio and precision ratio respectively. Following SC3D\cite{Siam3D2019}, their area-of-curve(AUC) is treated as the final metrics, denoted as ``Success'' and ``Precision'' . The larger the two metrics, the better this tracker is.

\subsection{Implementation Details}
\textbf{Network architecture.}
For both P2B and BAT, three set-abstraction (SA) layers of PointNet++ are utilized to extract high-level semantic features of key points. Its receptive radius is set to 0.3, 0.5, and 0.7 meters. To down-sample key points, the stride steps are 2, 4, and 8. And the dimension $D$ of is set to 256. For example, it can map the template of shape $N_T\times3$ to a tensor of shape $\frac{N_T}{8}\times256$ (\ie, $M_t=\frac{N_T}{8}$). For information embedding, P2B uses the target-specific feature augmentation (TFA) module to obtain the fused features $Z$ when BAT leverages a box-aware feature fusion (BAFF) module. The number of proposals, $N_{\mathcal{B}}$, is set to 64 by following the configuration of P2B and BAT. 
During feature decorrelation phase, we sample 5 basis functions $\{u_i\}_{t=1}^{N_u=5}$ from RKHS $\mathcal{H}_{RFF}$. Let $N(0,1)$ represent Gaussian distribution and $U(0,2\pi)$ uniform distribution, the function space $\mathcal{H}_{RFF}$ can be written as
\begin{equation}
    \begin{aligned}
        \mathcal{H}_{RFF}=\big\{&f(x)=\sqrt{2}\cos{(ax+b)}|\\&a\sim N(0,1), b\sim U(0,2\pi)\big\}. 
    \end{aligned}
\end{equation}
Similarly, we can also sample basis functions from $\mathcal{H}_{RFF}$ to obtain $\{v_t\}_{t=1}^{N_v=5}$. 

\renewcommand{\arraystretch}{1.5} 
\begin{table}[t]
  \centering
  \caption{performance on unseen categories of KITTI. The state-of-the-art methods are trained on two different modes: class-specific tracking and class-agnostic tracking. The context in the parentheses is unseen category. And the suffix ``-U'' represents the tracker that is trained under class-agnostic mode. }
  \label{tab:kit_unseen}
  \resizebox{1.0\linewidth}{!}{
      \begin{tabular}{c|c|cccc}
        \toprule
           \multirow{2}{*}{ Tracking Mode} & \multirow{2}{*}{ Methods} & \multicolumn{2}{c}{\makecell{KITTI Setting-1 \\ (Car)}} & \multicolumn{2}{c}{\makecell{KITTI Setting-2 \\ (Pedestrian)}}\\
            \cmidrule(lr){3-4} \cmidrule(lr){5-6}
            \quad & \quad & Success(\%) & Precision(\%) & Success(\%) & Precision(\%) \\
        \midrule
            \multirow{4}{*}{ Class-specific}& P2B\cite{P2B}  & 56.2 & 72.8 & 28.7 & 49.6\\
                                            & BAT\cite{BAT}  & 65.4 & 78.9 & 45.7 & 74.5\\
                                            & LTTR\cite{LTTR}& 65.0 & 77.1 & 33.2 & 56.8\\
                                            & V2B \cite{V2B} & 70.5 & 81.3 & 48.3 & 73.5\\
                                            & DSDM \cite{DSDM}& 65.1 & 75.8 & 32.5 & 51.4\\
        \midrule
            \multirow{2}{*}{ Class-agnostic}& P2B-U & 32.4 & 41.0 & 24.2 & 45.4 \\
                               & BAT-U & 24.9 & 33.7 & 12.9 & 21.8\\
        \bottomrule
      \end{tabular}
  }
\end{table}
\textbf{Training.}
For class-agnostic tracking, there are a total of four settings: KITTI Setting-1, KITTI Setting-2, NuScenes Setting-1, and NuScenes Setting-2. On each setting, the instances from the observed categories are only used to train the models. We create the template and search region from every instance tracklets. Specifically, for each frame, we add Gaussian perturbation to ground-truth bounding box, yielding a new bounding box. Centered on this box, we crop the search region with an offset of 2 meters. The template is then generated by aggregating the points inside the ground-truth boxes of the first and previous frames. Before sending these two sets of point clouds (template and search region) into the network, we transform them into a local coordinate system according to the corresponding bounding box. During training, a batch of such paired samples is selected randomly from the hybrid set consisting of the observed categories. In addition, when optimizing the Eq. \ref{eq:w}, we adopt SGD algorithm to solve the weights $\mathbf{w}$ with an iteration number of 20. Its learning rate is set to 1.0 and decayed 10 every 10 iterations. Regarding other hyper-parameters, we simply follow the baselines. Our experiments run on NVIDIA RTX-2080 GPUs with the help of pyTorch 1.7 and CUDA 10.1.

\textbf{Testing.}
We use well-trained models to track both observed and unseen objects. The template is updated by appending the tracked target into the initial one of the first frame. Considering the small motion increment between two consecutive frames, the search region is cropped around the tracking result of the previous frame. Both the template and search region are required to be transformed into their box coordinates system. Because the feature decorrelation method is committed to class-agnostic tracking by ameliorating the offline training scheme, it is not carried out during the testing phase. In other words, it will not affect the efficiency of the baselines during online tracking.

\renewcommand{\arraystretch}{1.5} 
\begin{table}[t]
  \centering
  \caption{performance on observed categories of KITTI. The class-specific trackers train different models for different categories. We report the mean value of success and precision over all categories. The class-agnostic trackers use our setting to train only one model for all categories.}
  \label{tab:kit_observe}
  \resizebox{1.0\linewidth}{!}{
      \begin{tabular}{c|c|cccc}
        \toprule
           \multirow{2}{*}{ Tracking Mode} & \multirow{2}{*}{Methods} & \multicolumn{2}{c}{\makecell{KITTI Setting-1 \\ (Pedestrian, Van, Cyclist)}} & \multicolumn{2}{c}{\makecell{KITTI Setting-2 \\(Car, Cyclist, Van)}}\\
            \cmidrule(lr){3-4} \cmidrule(lr){5-6}
            \quad & \quad & Success(\%) & Precision(\%) & Success(\%) & Precision(\%) \\
        \midrule
            \multirow{4}{*}{ Class-specific} & P2B\cite{P2B}  & 30.8 & 49.2 & 52.9 & 67.9\\
                                    & BAT\cite{BAT}  & 46.3 & 72.1 & 62.1 & 75.7\\
                                    & LTTR\cite{LTTR}& 35.0 & 56.3 & 60.5 & 72.7\\
                                    & V2B \cite{V2B} & 48.3 & 70.1 & 66.2 & 76.4\\
                                    & DSDM \cite{DSDM} & 37.0 & 53.9 & 63.1 & 73.4\\
        \midrule
            \multirow{2}{*}{ Class-agnostic} & P2B-U & 38.6 & 57.4 & 55.6 & 70.1\\
                                             & BAT-U & 33.9 & 49.8 & 59.6 & 73.1\\
        \bottomrule
      \end{tabular}
  }
\end{table}

\renewcommand{\arraystretch}{1.5} 
\begin{table*}[t]
  \centering
  \caption{The experimental settings on KITTI and NuScenes.
  The suffix ``-W'' indicates it uses the feature decorrelation method, "-S" uses Sharpness-Aware Minimization, and "-U" represents a plain version.}
  \label{tab:kit_nus}
      \begin{tabular}{c|c|cccccccc}
        \toprule
           \multirow{2}{*}{Metrics } & \multirow{2}{*}{ Methods} & \multicolumn{2}{c}{KITTI Setting-1} & \multicolumn{2}{c}{KITTI Setting-2} & \multicolumn{2}{c}{NuScenes Setting-1} & \multicolumn{2}{c}{NuScenes Setting-2}\\
            \cmidrule(lr){3-4} \cmidrule(lr){5-6} \cmidrule(lr){7-8} \cmidrule(lr){9-10}
            \quad & \quad & \makecell{Observed} & \makecell{\textbf{Unseen}} & \makecell{Observed} & \makecell{\textbf{Unseen}} & \makecell{Observed} & \makecell{\textbf{Unseen}} & \makecell{Observed} & \makecell{\textbf{Unseen}} \\
        \midrule
            \multirow{6}{*}{Success(\%)}& P2B-U & 38.6 & 32.4 & 55.6 & 24.2 & 31.1 & 34.3 & 42.5 & 14.6 \\
                                    & P2B-W & 41.0 & \textbf{34.7} & 56.2 & \textbf{26.6} & 32.6 & \textbf{37.4} & 41.4 & 16.2\\
                                    & P2B-S & 31.4 & 21.3 & 46.9 & 20.9 & 26.1 & 29.7 & 36.5 & 13.3\\
                                    & BAT-U & 33.9 & 24.9 & 59.6 & 12.9 & 30.9 & 33.6 & 38.6 & 16.9 \\
                                    & BAT-W & 35.7 & 28.1 & 61.8 & 14.6 & 33.4 & 36.0 & 41.9 & 19.0\\
                                    & BAT-S & 22.5 & 18.1 & 43.5 & 10.6 & 29.7 & 32.8 & 38.5 & \textbf{19.3}\\
        \midrule
        \midrule
            \multirow{6}{*}{Precision(\%)}& P2B-U & 57.4 & 41.0 & 70.1 & 45.4 & 46.1 & 35.3 & 45.9 & 33.7 \\
                                      & P2B-W & 61.5 & \textbf{45.6} & 70.3 & \textbf{47.8} & 48.3 & \textbf{39.8} & 44.0 & 35.8\\
                                      & P2B-S & 51.7 & 28.2 & 62.6 & 41.4 & 41.3 & 30.8 & 39.7 & 32.2\\
                                      & BAT-U & 49.8 & 33.7 & 73.1 & 21.8 & 44.0 & 35.4 & 40.5 & 37.1 \\
                                      & BAT-W & 50.3 & 37.3 & 74.5 & 27.9 & 47.8 & 37.4 & 43.8 & 38.1\\
                                      & BAT-S & 38.0 & 24.2 & 58.3 & 18.8 & 45.6 & 35.6 & 41.1 & \textbf{40.5}\\
        \bottomrule
      \end{tabular}
\end{table*}

\subsection{Preliminary Experiments}
Since P2B\cite{P2B} was proposed for 3D SOT, there have sprout many improved algorithms, including LTTR\cite{LTTR}, V2B \cite{V2B}, DSDM\cite{DSDM} and BAT\cite{BAT}. These methods have good performance under the class-specific tracking mode. 
Based on them, we conduct some preliminary experiments, which is used to show the performance change when current methods are shifted from class-specific to class-agnostic tracking mode.
Specifically, taking P2B and BAT as the baselines, we retrain them use the class-agnostic mode mentioned in Section \ref{sec:setting}, and the resulting models are denoted as P2B-U and BAT-U. We compare them on both observed and unseen categories.

Table \ref{tab:kit_unseen} reports the performance of P2B-U and BAT-U on unseen categories of KITTI Setting-1 and Setting-2, and compare them with methods using class-specific tracking mode.
Generally, class-specific trackers achieve high performance because the samples of training and testing sets are derived from the same category. For example, according to the success/precision metrics, P2B reaches 56.2\% /72.8\% on cars and 28.7\%/49.6\% on pedestrians. Its improved versions like BAT and LTTR can further achieve significant improvements. In particular, BAT surpasses P2B by 9.2\%/6.1\% on cars and by 17\%/23.9\% on pedestrians. However, despite these improvements, they still have a long path towards class-agnostic tracking. From the last two rows of Table \ref{tab:kit_unseen}, P2B-U only obtains the performance of 32.4\%/41.0\% on cars, which significantly decreases by 13.8\%/31.8\% compared with P2B. BAT-U and BAT even differ by 30.5\%/45.2\% on cars. A similar phenomenon also occurred on pedestrians. In light of this, it is very necessary and meaningful to study some methods to promote the development of class-agnostic tracking. Moreover, there is another finding that BAT-U gets worse performance than P2B-U although its original version, BAT, is greatly superior to P2B. The reason is that it may confuse the network if the Box-Cloud distributions are regressed forcibly on all classes without any prior. This illustrates that the accuracy of class-agnostic tracking is not positively correlated with that of the original class-specific tracker.

While unseen categories can evaluate the generalization ability, we still expect that the trackers can remain original accuracy on observed categories. Therefore, Table \ref{tab:kit_observe} summarizes the tracking results on observed categories of KITTI Setting-1 and Setting-2. 
Herein, the class-agnostic trackers learn a general model using mixed samples of different categories, while the class-specific trackers separately train multiple models for different categories. 
As shown in this table, class-specific trackers, BAT and V2B, win the top two places.
A more interesting finding is that P2B-U and BAT-U have opposite performance changes after training on KITTI Setting-1 and Setting-2. Specifically, when tracking the objects from pedestrians, vans, and cyclists (Setting-1), P2B-U brings the improvement of 7.8\%/8.2\% in terms of success/precision metrics compared to its original version P2B, but BAT-U decreases by 12.4\%/6.5\% compared to BAT. Similarly, when evaluating on the Setting-2, the gain of P2B-U is 2.7\%/2.2\%, but BAT-U is reduced by 2.5\%/2.6\%. The reason may lie in that the information embedding of BAT is sensitive when training multiple categories together. This stimulates our curiosity to explore the effects of experimental configurations by mixing different categories. Analysis on this part will be shown in the next section \ref{sec:config}.

To sum up, according to the above preliminary experiments, the recently advanced trackers still require great efforts to simultaneously obtain accurate tracking results on unseen and observed classes, which validates the standpoints that we put forward in the introduction. Next, we will verify the proposed feature decorrelation method for class-agnostic tracking. 

\begin{figure*}[thb]
    \centering
    \includegraphics[width=0.95\linewidth]{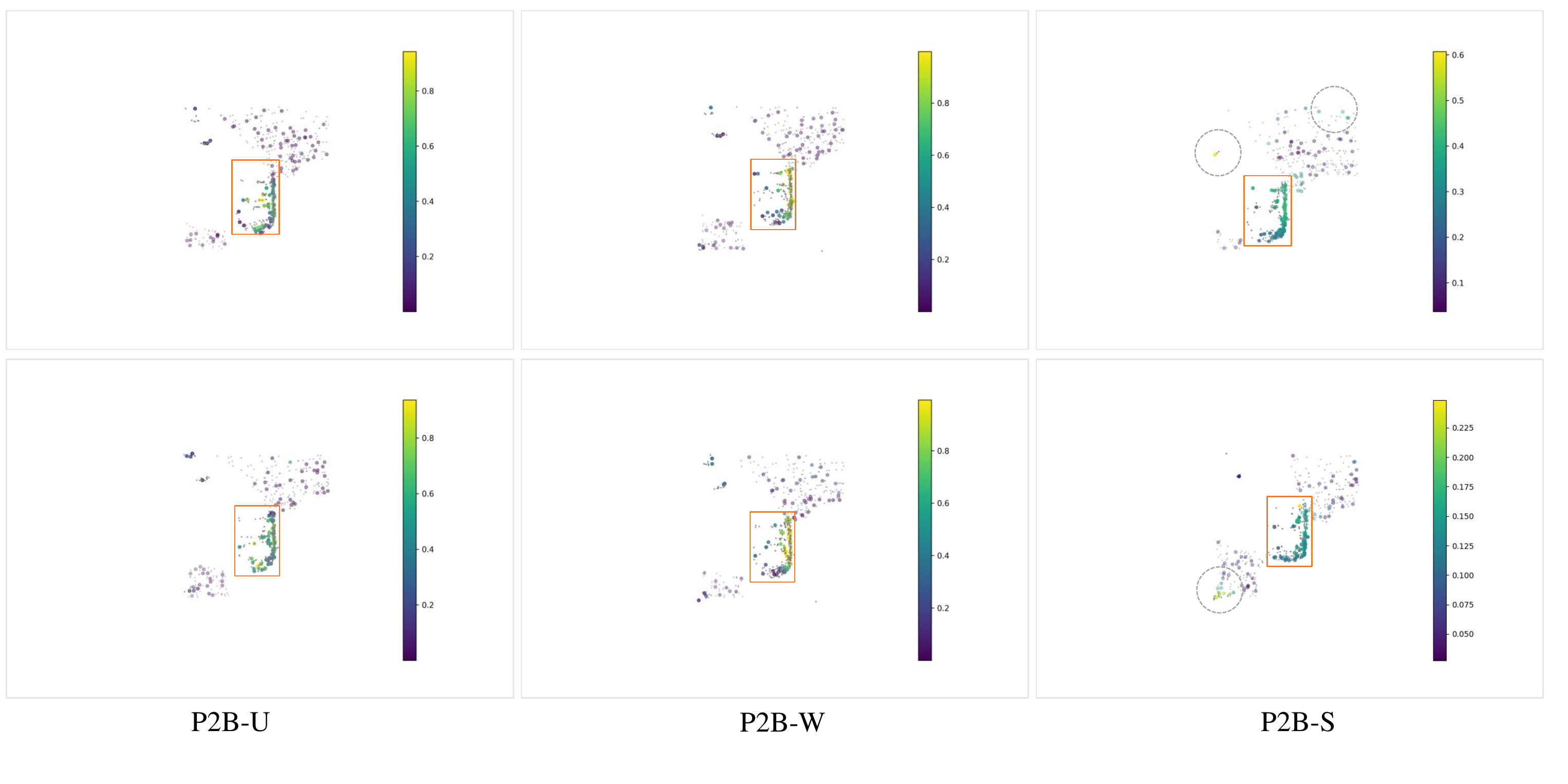}
    \caption{P2B visualization of the fused features for search region. The targets are covered by orange rectangle. The points highlighted by gray circles mean that the corresponding method produces spurious features for them. The color bar represents how well the current point matches the target. The more yellow the point is, the greater the probability that the point lies on the target surface.}
    \label{fig:p2b}
\end{figure*}

\begin{figure*}[th]
    \centering
    \includegraphics[width=0.95\linewidth]{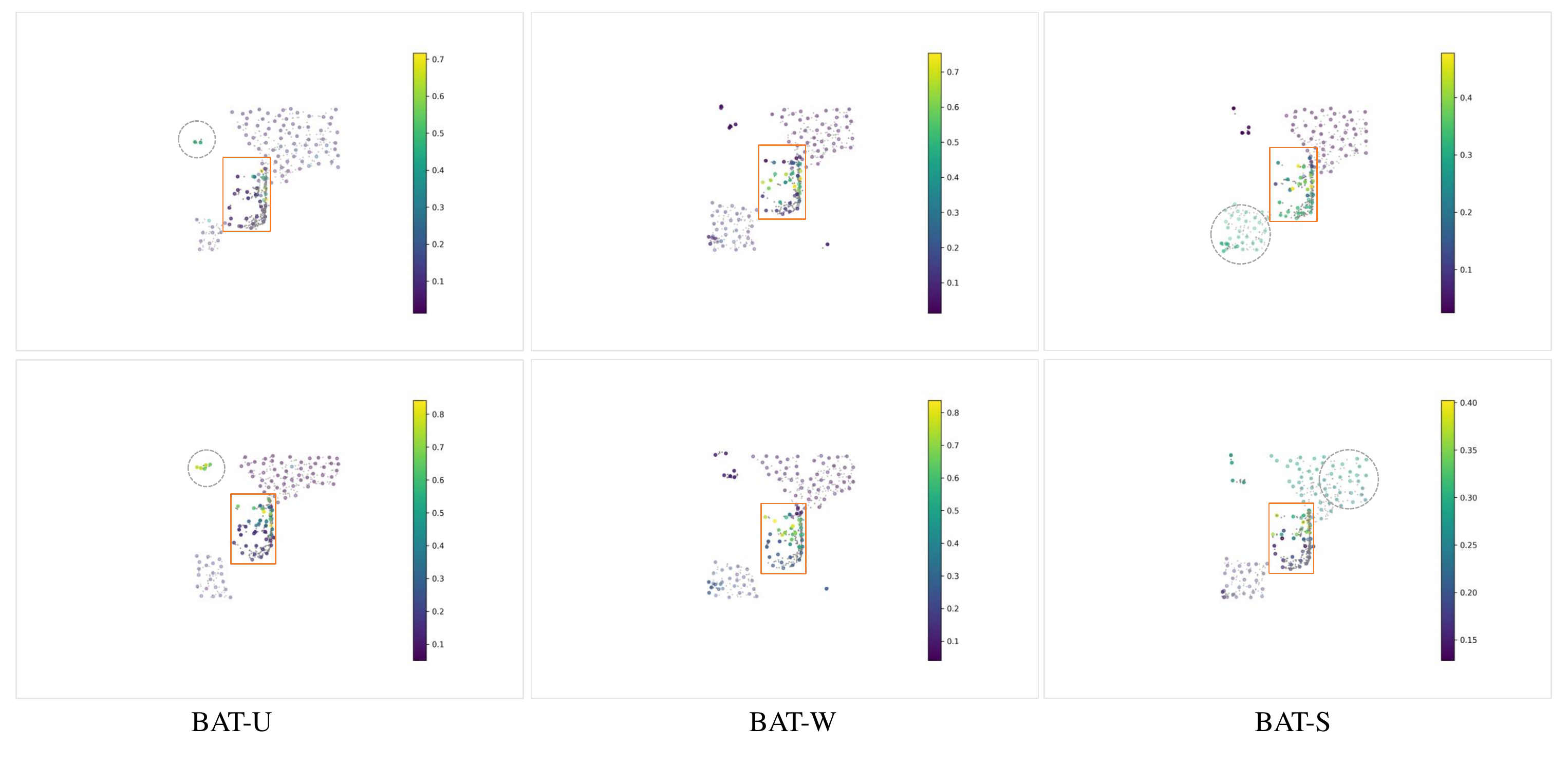}
    \caption{BAT visualization of the fused features for search region. The targets are covered by orange rectangle. The points highlighted by gray circles mean that the corresponding method produces spurious features for them. The color bar represents how well the current point matches the target. The more yellow the point is, the greater the probability that the point lies on the target surface.}
    \label{fig:bat}
\end{figure*}

\begin{figure}[th]
    \centering
    \includegraphics[width=0.95\linewidth]{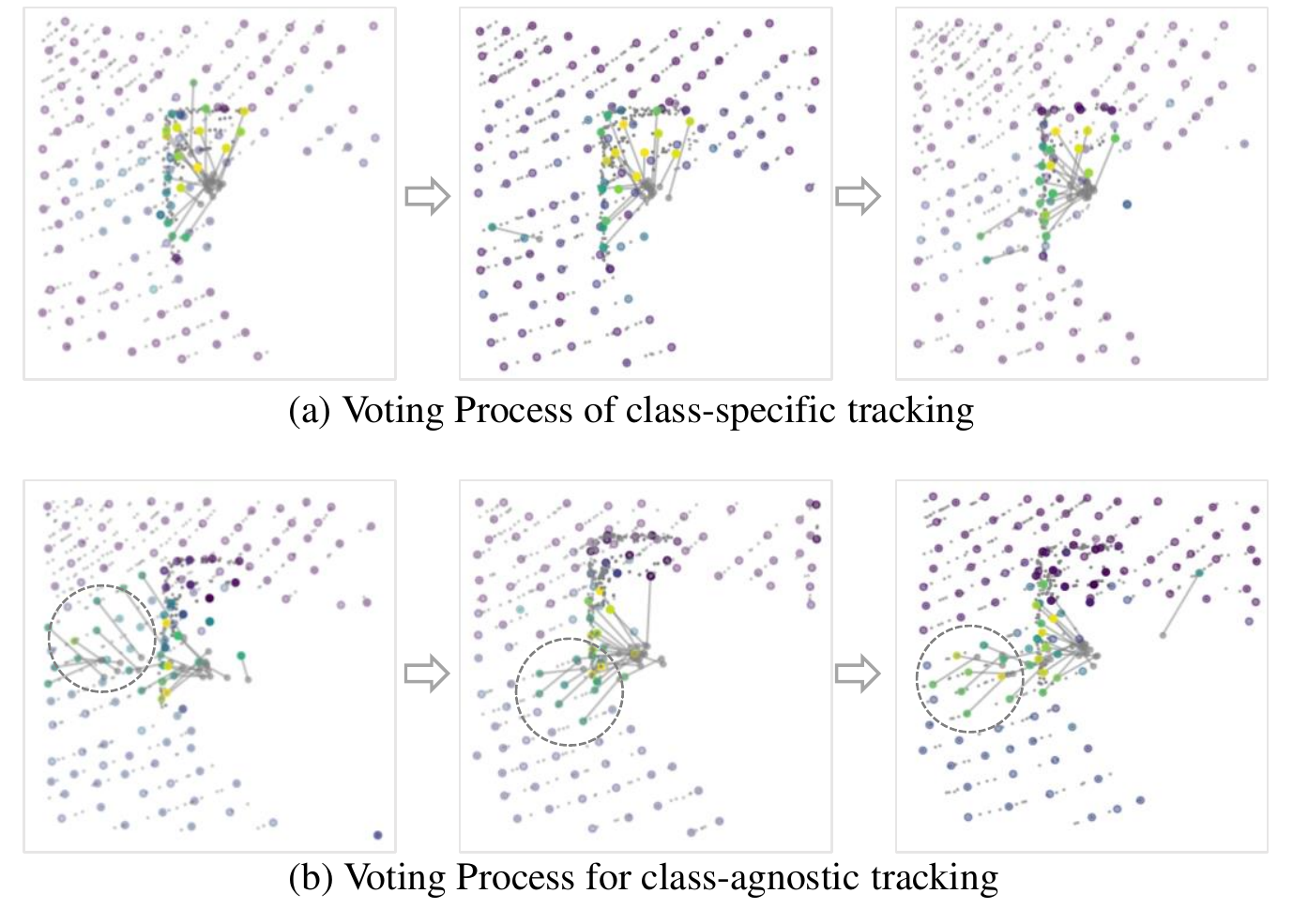}
    \caption{Impact of the spurious correlation on class-specific and class-agnostic tracking. The voting process for the location prediction is plotted. (a) shows the results of class-specific tracking, where the irrelevant features only have slight influence. But, (b) shows the results of class-agnostic tracking, which is obviously interfered by the irrelevant features, \ie, background. We highlight it by gray circles.}
    \label{fig:background}
\end{figure}
\subsection{Comparison with Baselines}
In this subsection, we validate the effectiveness of the proposed module by comparing it with P2B-U and BAT-U. For clarity, we use the suffix ``-W'' to represent the variations that equip with the feature decorrelation method proposed in Section \ref{sec:sw}. Under our new setup, all trackers are supposed to be tested on both unseen and observed categories. To comprehensively compare with different methods, we also consider another large-scale dataset NuScenes in addition to KITTI.

\textbf{Results on KITTI.}
As reported in Table \ref{tab:kit_nus}, when tracking unseen classes on KITTI, P2B-W achieves the best performance. When tracking observed classes, BAT-W ranks first on Setting-2 and P2B-W wins first place on setting-1. Concretely, compared with P2B-U in success/precision metrics, P2B-W can track unseen shapes with the considerable improvements of 2.3\%/4.6\% on Setting-1 and 2.4\%/2.4\% on Setting-2; and even on observed shapes, it can reap the rewards of 2.4\%/4.1\% on Setting-1. Similarly, for unseen shapes, BAT-W outperforms BAT-U by 3.2\%/3.6\% on Setting-1 and 1.7\%/6.1\% on Setting-2. As for observed shapes, BAT-W can still obtain 35.7\%/50.3\% and 61.8\%/74.5\% performance on Setting-1 and Setting-2 respectively, which are 1.8\%/0.5\% and 2.2\%/1.4\% higher than BAT-U. This demonstrates that the proposed module plays an important role in class-agnostic tracking.  From another perspective, it is worth mentioning that P2B-W overpasses BAT-W by 6.6\%/8.3\% and 12\%/19.9\% on two unseen settings although BAT is better than P2B in class-specific tracking, which tells us strong baselines may have inconsistent performance in class-agnostic tracking. 

\textbf{Results on NuScenes.}
In the last four columns of Table \ref{tab:kit_nus}, we report the tracking results on NuScenes Setting-1 and Setting-2. Overall, P2B-W and BAT-W achieve the best results on unseen categories. In particular, compared with P2B-U, P2B-W increases by 3.1\%/4.5\% and 1.6\%/2.1\% on Setting-1 and Setting-2, respectively. And BAT-W also achieves the improvements of 2.4\%/2.0\% and 2.1\%/1.0\% in comparison to its baseline BAT-U. These results further illustrate the effectiveness of the feature decorrelation based on sample weighting. As for observed categories, both methods with suffix ``-W'' obtains competitive performance against their baselines. For example, on setting-1, BAT-U is enhanced from 30.9\%/44.0\% to 33.4\%/47.8\% in success/precision metrics, while P2B-U from 31.1\%/46.1\% to 32.6\%/48.3\%.  

From a whole point of view, we further draw the following conclusions. Firstly, according to the overall performance, the variations of P2B family are better than BAT family. Secondly, even though our proposed method is demonstrated to be conductive to class-agnostic tracking, there is still potential room to catch up with the performance of class-specific tracking. Lastly, similar shapes may provide some useful prior for tracking unseen categories. For instance, the distribution of vans is similar to that of cars on KITTI and the distribution of trucks resembles that of cars on NuScenes, which generally brings about higher results when cars are unseen class on Setting-1.

\subsection{Other Method for Class-agnostic Tracking}
Sharpness-Aware Minimization (SAM)\cite{sam} recently has attracted increasing attention in model generalization. The theoretical support behind this method is that when optimizing the non-convex loss function of the deep neural network, the parameters under flat extreme usually have more generalization ability than those under sharp extreme. Therefore, SAM not only pursues the minimum loss value but also considers the flatness of the loss landscape, where the flatness means that a $\delta$-neighborhood of the solved parameters ought to have very approximate loss values. In this work, this SAM method also is taken into account so as to deal with the class-agnostic tracking problem. Following the previous framework, we apply it into the two baselines---P2B and BAT---under our proposed settings. The suffix ``-S'' is utilized to denote the corresponding variations trained with SAM. For well-round comparisons, we evaluate them on both KITTI and NuScenes.

As we can see from Table \ref{tab:kit_nus}, the methods with the suffix ``-S'' do not have stable performance on unseen categories. For example,  BAT-S achieves a satisfactory performance solely on NuScenes Setting-2, while all others are inferior to P2B-W and BAT-W. Additionally, for observed categories, P2B-S and BAT-S obtain worse tracking results than P2B-W and BAT-W respectively. Especially, on KITTI Setting-2, BAT-W outperforms BAT-S in success/precision metrics, up to 18.3\%/16.2\%. Therefore, the robustness of such variants is a key problem worth considering for class-agnostic tracking. We conjecture that the solution of SAM may approach the edge of flatness, and gathering stochastic weights densely may be able to explore more flatter extreme \cite{swad}.

\subsection{Visualization}
\textbf{Phenomenon of the spurious correlation.} One may think that the class-specific tracking also involves the spurious correlation from irrelevant features. But in fact, the class-specific tracking is to train and test a model on the same class, which can handle the observed distribution of a category well. When confronted with unseen classes (\ie, class-agnostic tracking), the model becomes confused and the impact of irrelevant features will be amplified. To validate this, we show the visualization results of the subsequent voting process in Figure \ref{fig:background}. As we can seen, for the same instance, the class-specific tracking is slightly affected by the irrelevant features, whereas  
and the class-specific tracking even clusters the background points together. 

\textbf{Effectiveness for the fused features.}
As mentioned before, we aim at making the fused features consistent among foreground points while being able to distinguish background points. To verify whether this goal is achieved, we visualize semantic information based on the fused features. Figure \ref{fig:p2b} shows the visualization results of P2B-U, P2B-W, and P2B-S. Figure \ref{fig:bat} plots the results of BAT-U, BAT-W, and BAT-S. All the models are evaluated on unseen categories. In these figures, the more yellow the point is, the greater the probability that the point lies on the target surface. We cover the targets with the orange rectangles. From Figure \ref{fig:p2b}, we can observe that P2B-W has a much brighter yellow on the scanned points of the target compared with P2B-U and P2B-S. In particular, P2B-S even generates a high similarity for those features derived from background points (labeled in circles), which will mislead the subsequent tracking procedure. In addition, as shown in Figure \ref{fig:bat}, BAT-W can produce more discriminative cases where the foreground points tend to be yellow and the background points are dark consistently. Note that, both BAT-U and BAT-S have the phenomenon of outputting spurious features for background points (labeled in circles).

\renewcommand{\arraystretch}{1.5} 
\begin{table*}[t]
  \centering
  \caption{Effectiveness of using mixed samples of different categories. Instead of training the model on a single category, we investigate the expansibility of trackers via putting different categories together. The results are tested on this observed categories.}
  \label{tab:mixed_samples}
      \begin{tabular}{c|c|cccc}
        \toprule
           \multirow{2}{*}{ Datasets} & \multirow{2}{*}{ Configurations} & \multicolumn{2}{c}{P2B-W} & \multicolumn{2}{c}{BAT-W}\\
            \cmidrule(lr){3-4} \cmidrule(lr){5-6}
            \quad & \quad& Success(\%) & Precision(\%) & Success(\%) & Precision(\%)  \\
        \midrule
            \multirow{3}{*}{ KITTI} & Car, Cyclist, Van       & 56.2 & 70.3 & 61.8 & 74.5\\
                               & Pedestrian, Van, Cyclist     & 41.0 & 61.5 & 35.7 & 50.3\\
                               & Car, Pedestrian,Van, Cyclist & 48.9 & 67.1 & 42.4 & 58.5\\
        \midrule
            \multirow{3}{*}{ NuScenes} & Car, Bicycle, Truck     & 41.4 & 44.0 & 41.9 & 43.8\\
                               & Pedestrian, Bicycle, Truck      & 32.6 & 48.3 & 33.4 & 47.8\\
                               & Car, Pedestrian, Bicycle, Truck & 35.7 & 42.4 & 34.9 & 41.2\\
        \bottomrule
      \end{tabular}
\end{table*}
\subsection{Discussions}\label{sec:config}
\textbf{What kind of baselines can work well for class-agnostic tracking? }
Like 2D visual tracking, the ultimate goal of 3D SOT is to train a general model capable of tracking any specified target no matter whether it is observed or not during training. Therefore, it is inevitable to offline train a model using mixed samples from different categories. Here, we discuss the impact of experimental configurations that mix different categories. On both KITTI and NuScenes, we run three different combinations and test the trained models on the same categories. As reported in Table \ref{tab:mixed_samples}, we observe that P2B-W has a better expansibility than BAT-w in all configurations. We deem that a good 3D tracker should balance the two aspects well. Firstly, for intra-class objects, the point clouds scanned by LiDAR from a single view usually have the same patterns, a tracker thus should be able to discern the target from these inter-class distractors. Secondly, for inter-class objects, the distribution and size often have obvious discrepancies, which requires a tracker to be robust to the shift of inter-class objects. P2B and BAT overlook the first and second points, respectively. In particular, BAT is easy to confuse the network if the Box- Cloud is forced to be regressed on all kinds of classes without any prior. 

\textbf{Limitations and Future works.}
The sample weighting method is originally designed for out-of-distribution generalization by balancing each variable in each sample. In this work, we follow the literature \cite{sw_image} to iteratively adjust the training instances in a mini-batch. It can be explored to adopt other strategies to update global weights. In addition, there are several other ways to be worth studying for class-agnostic tracking apart from sample weighting. For instance, one of the solutions is that the network can learn to learn from the shift between observed and unseen domains. Therefore, we can let meta-learning\cite{MAML} simulate the ``learning-then-generalization" procedure, and then apply it to online tracking.
In the realm of 2D visual tracking, some researchers \cite{TID2020, BridgingDT2019} treated each video clip as one domain, and leveraged meta-learning to learn fast adaptation ability. However, the characteristic of point cloud data is that the obvious difference between categories is only reflected in the spatial distribution. When each point cloud sequence is treated as a domain, especially when the number of categories is unbalanced, it cannot give full play to its advantages. Alternatively, taking inspiration from \cite{generalReid, Person30K}, we can view each scanned category as one domain to design class-agnostic tracking. 
Another solution is that we can use data generation to enhance the generalization in real-world scenarios. For example, the deformed point cloud learned from adversarial attack \cite{3DVField} can be utilized to expand the original data distribution. And based on the idea of PointAugment \cite{pointaugment}, we can even control the generation of more diverse samples during training.


\section{Conclusion}
This work investigates the class-agnostic tracking task in 3D LiDAR point clouds. In order to promote the development of this task, we construct the observed and unseen categories for training and testing phases, based on the two large-scale datasets KITTI and NuScenes. Furthermore, considering that this task involves the domain shift problem when tacking unseen categories, we propose a feature decorrelation module (FDM), which leverages the sample weighting method to make the features produced by information embedding more discriminative. Experiments on KITTI and NuScens have demonstrated that the baselines P2B and BAT can achieve considerable improvements after equipping with the FDM. In addition, we also provide some insights based on the experimental analysis for future studying of class-agnostic tracking.



\ifCLASSOPTIONcaptionsoff
  \newpage
\fi



\bibliographystyle{IEEEtran}
\bibliography{mybib}

\begin{thebibliography}{10}
\providecommand{\url}[1]{#1}
\csname url@samestyle\endcsname
\providecommand{\newblock}{\relax}
\providecommand{\bibinfo}[2]{#2}
\providecommand{\BIBentrySTDinterwordspacing}{\spaceskip=0pt\relax}
\providecommand{\BIBentryALTinterwordstretchfactor}{4}
\providecommand{\BIBentryALTinterwordspacing}{\spaceskip=\fontdimen2\font plus
\BIBentryALTinterwordstretchfactor\fontdimen3\font minus
  \fontdimen4\font\relax}
\providecommand{\BIBforeignlanguage}[2]{{%
\expandafter\ifx\csname l@#1\endcsname\relax
\typeout{** WARNING: IEEEtran.bst: No hyphenation pattern has been}%
\typeout{** loaded for the language `#1'. Using the pattern for}%
\typeout{** the default language instead.}%
\else
\language=\csname l@#1\endcsname
\fi
#2}}
\providecommand{\BIBdecl}{\relax}
\BIBdecl

\bibitem{Argoverse2019}
M.-F. Chang, J.~Lambert, P.~Sangkloy, J.~Singh, S.~Bak, A.~T. Hartnett,
  D.~Wang, P.~Carr, S.~Lucey, D.~Ramanan, and J.~Hays, ``Argoverse: {3D}
  tracking and forecasting with rich maps,'' in \emph{CVPR}, 2019.

\bibitem{FaF2018}
W.~Luo, B.~Yang, and R.~Urtasun, ``Fast and {Furious}: Real time end-to-end
  {3D} detection, tracking and motion forecasting with a single convolutional
  net,'' in \emph{CVPR}, 2018.

\bibitem{Wang2017RobotsNavi}
M.~Wang, D.~Su, L.~Shi, Y.~Liu, and J.~V. Mir{\'o}, ``Real-time {3D} human
  tracking for mobile robots with multisensors,'' in \emph{ICRA}, 2017.

\bibitem{annotation}
B.~{Wang}, V.~{Wu}, B.~{Wu}, and K.~{Keutzer}, ``{LATTE}: Accelerating {LiDAR}
  point cloud annotation via sensor fusion, one-click annotation, and
  tracking,'' in \emph{IEEE Intelligent Transportation Systems Conference
  (ITSC)}, 2019, pp. 265--272.

\bibitem{MultiStream2020TIP}
K.~{Li}, Y.~{Kong}, and Y.~{Fu}, ``Visual object tracking via multi-stream deep
  similarity learning networks,'' \emph{IEEE Trans. Image Process.}, vol.~29,
  pp. 3311--3320, 2020.

\bibitem{PointNet2017}
C.~R. Qi, H.~Su, K.~Mo, and L.~J. Guibas, ``{PointNet}: Deep learning on point
  sets for {3D} classification and segmentation,'' in \emph{CVPR}, 2017.

\bibitem{PointNetPP2017}
C.~R. Qi, L.~Yi, H.~Su, and L.~J. Guibas, ``{PointNet++}: Deep hierarchical
  feature learning on point sets in a metric space,'' in \emph{NeurIPS}, 2017.

\bibitem{Siam3D2019}
S.~Giancola, J.~Zarzar, and B.~Ghanem, ``Leveraging shape completion for {3D}
  {Siamese} tracking,'' in \emph{CVPR}, 2019.

\bibitem{P2B}
H.~Qi, C.~Feng, Z.~Cao, F.~Zhao, and Y.~Xiao, ``{P2B}: Point-to-box network for
  3d object tracking in point clouds,'' in \emph{CVPR}, 2020.

\bibitem{BAT}
C.~Zheng, X.~Yan, J.~Gao, W.~Zhao, W.~Zhang, Z.~Li, and S.~Cui, ``{Box-Aware}
  feature enhancement for single object tracking on point clouds,'' in
  \emph{ICCV}, 2021, pp. 13\,199--13\,208.

\bibitem{sint}
R.~Tao, E.~Gavves, and A.~W. Smeulders, ``Siamese instance search for
  tracking,'' in \emph{CVPR}, 2016, pp. 1420--1429.

\bibitem{KITTI2013}
A.~Geiger, P.~Lenz, C.~Stiller, and R.~Urtasun, ``Vision meets robotics: The
  {KITTI} dataset,'' \emph{International Journal of Robotics Research},
  vol.~32, no.~11, pp. 1231--1237, 2013.

\bibitem{nuscenes}
H.~Caesar, V.~Bankiti, A.~H. Lang, S.~Vora, V.~E. Liong, Q.~Xu, A.~Krishnan,
  Y.~Pan, G.~Baldan, and O.~Beijbom, ``nuscenes: A multimodal dataset for
  autonomous driving,'' in \emph{CVPR}, 2020, pp. 11\,621--11\,631.

\bibitem{sw_nature}
P.~Cui and S.~Athey, ``Stable learning establishes some common ground between
  causal inference and machine learning,'' \emph{Nature Machine Intelligence},
  vol.~4, pp. 110--115, 2022.

\bibitem{rrf}
A.~Rahimi, B.~Recht \emph{et~al.}, ``Random features for large-scale kernel
  machines.'' in \emph{Nuerips}, vol.~3, no.~4, 2007, p.~5.

\bibitem{Held2013PrecisionTW}
D.~Held, J.~Levinson, and S.~Thrun, ``Precision tracking with sparse {3D} and
  dense color {2D} data,'' in \emph{ICRA}, 2013.

\bibitem{MotionBased}
A.~Dewan, T.~Caselitz, G.~D. Tipaldi, and W.~Burgard, ``Motion-based detection
  and tracking in {3D} lidar scans,'' in \emph{ICRA}, 2015.

\bibitem{rgb_lidar2016}
A.~Asvadi, P.~Gir{\~a}o, P.~Peixoto, and U.~Nunes, ``3d object tracking using
  rgb and lidar data,'' in \emph{IEEE International Conference on Intelligent
  Transportation Systems (ITSC)}, 2016.

\bibitem{SOTracker}
Z.~Pang, Z.~Li, and N.~Wang, ``Model-free vehicle tracking and state estimation
  in point cloud sequences,'' in \emph{IROS}, 2021.

\bibitem{SiamFC2016}
L.~Bertinetto, J.~Valmadre, J.~F. Henriques, A.~Vedaldi, and P.~H.~S. Torr,
  ``Fully-convolutional {Siamese} networks for object tracking,'' in
  \emph{ECCV}, 2016.

\bibitem{retracker2020}
T.~Feng, L.~Jiao, H.~Zhu, and L.~Sun, ``A novel object re-track framework for
  3d point clouds,'' in \emph{ACM International Conference on Multimedia},
  2020, p. 3118–3126.

\bibitem{LearningIW2021}
S.~Tian, X.~Liu, M.~Liu, Y.~Bian, J.~Gao, and B.~Yin, ``Learning the
  incremental warp for 3d vehicle tracking in lidar point clouds,''
  \emph{Remote Sensing}, vol.~13, p. 2770, 2021.

\bibitem{DeepHough2019}
C.~R. Qi, O.~Litany, K.~He, and L.~J. Guibas, ``Deep hough voting for {3D}
  object detection in point clouds,'' in \emph{ICCV}, 2019.

\bibitem{SiamRPN2018}
B.~Li, J.~Yan, W.~Wu, Z.~Zhu, and X.~Hu, ``High performance visual tracking
  with {Siamese} region proposal network,'' in \emph{CVPR}, 2018.

\bibitem{DSDM}
S.~Tian, B.~Liu, H.~Tan, J.~Liu, M.~Liu, and X.~Liu, ``Deep supervised descent
  method with multiple seeds generation for {3D} tracking in point cloud,''
  \emph{IEEE Transactions on Industrial Informatics}, pp. 1--1, 2021.

\bibitem{V2B}
L.~Hui, L.~Wang, M.~Cheng, J.~Xie, and J.~Yang, ``3d siamese voxel-to-bev
  tracker for sparse point clouds,'' in \emph{Advances in Neural Information
  Processing Systems}, 2021.

\bibitem{LTTR}
Y.~Cui, Z.~Fang, J.~Shan, Z.~Gu, and S.~Zhou, ``3d object tracking with
  transformer,'' in \emph{BMVC}, 2021.

\bibitem{PTT2021}
J.~Shan, S.~Zhou, Z.~Fang, and Y.~Cui, ``{PTT}: Point-track-transformer module
  for 3d single object tracking in point clouds,'' in \emph{IROS}, 2021, p.
  8133–8139.

\bibitem{pointnetLK2019}
Y.~Aoki, H.~Goforth, R.~A. Srivatsan, and S.~Lucey, ``{PointNetLK}: Robust \&
  efficient point cloud registration using {PointNet},'' in \emph{CVPR}, 2019.

\bibitem{dcp}
Y.~Wang and J.~M. Solomon, ``Deep closest point: Learning representations for
  point cloud registration,'' in \emph{Proceedings of the IEEE/CVF
  International Conference on Computer Vision}, 2019, pp. 3523--3532.

\bibitem{dgr}
C.~Choy, W.~Dong, and V.~Koltun, ``Deep global registration,'' in
  \emph{Proceedings of the IEEE/CVF conference on computer vision and pattern
  recognition}, 2020, pp. 2514--2523.

\bibitem{predator}
S.~Huang, Z.~Gojcic, M.~Usvyatsov, A.~Wieser, and K.~Schindler, ``Predator:
  Registration of 3d point clouds with low overlap,'' in \emph{Proceedings of
  the IEEE/CVF Conference on Computer Vision and Pattern Recognition}, 2021,
  pp. 4267--4276.

\bibitem{alignnet}
J.~Gro{\ss}, A.~O{\v{s}}ep, and B.~Leibe, ``Alignnet-3d: Fast point cloud
  registration of partially observed objects,'' in \emph{2019 International
  Conference on 3D Vision (3DV)}, 2019, pp. 623--632.

\bibitem{pointnetLK_revisit}
X.~Li, J.~K. Pontes, and S.~Lucey, ``Pointnetlk revisited,'' in \emph{CVPR},
  June 2021, pp. 12\,763--12\,772.

\bibitem{sw_kuang}
K.~Kuang, R.~Xiong, P.~Cui, S.~Athey, and B.~Li, ``Stable prediction with model
  misspecification and agnostic distribution shift,'' in \emph{AAAI}, 2020, p.
  4485–4492.

\bibitem{sw_shen}
Z.~Shen, P.~Cui, T.~Zhang, and K.~Kuang, ``Stable learning via sample
  reweighting,'' in \emph{AAAI}, 2020, p. 5692–5699.

\bibitem{sw_dvd}
Z.~Shen, P.~Cui, J.~Liu, T.~Zhang, B.~Li, and Z.~Chen, ``Stable learning via
  differentiated variable decorrelation,'' in \emph{Proceedings of the 26th ACM
  SIGKDD International Conference on Knowledge Discovery \& Data Mining}, 2020,
  pp. 2185--2193.

\bibitem{sw_nico}
Y.~He, Z.~Shen, and P.~Cui, ``Towards non-iid image classification: A dataset
  and baselines,'' \emph{Pattern Recognition}, vol. 110, p. 107383, 2021.

\bibitem{sw_image}
X.~Zhang, P.~Cui, R.~Xu, L.~Zhou, Y.~He, and Z.~Shen, ``Deep stable learning
  for out-of-distribution generalization,'' in \emph{CVPR}, 2021, pp.
  5372--5382.

\bibitem{3D-SiamRPN}
X.~Xiong, S.~Zhou, Y.~Cui, and S.~Scherer, ``{3D-SiamRPN}: An end-to-end
  learning method for real-time 3d single object tracking using raw point
  cloud,'' \emph{IEEE Sensors Journal}, vol.~21, no.~4, pp. 4995--5011, 2021.

\bibitem{OTB2015}
Y.~Wu, J.~Lim, and M.-H. Yang, ``Object tracking benchmark,'' \emph{IEEE
  Transactions on Pattern Analysis and MachineIntelligence}, vol.~37, no.~9,
  pp. 1834--1848, 2015.

\bibitem{sam}
P.~{Foret}, A.~{Kleiner}, H.~{Mobahi}, and B.~{Neyshabur}, ``Sharpness-aware
  minimization for efficiently improving generalization,'' in \emph{ICLR},
  2021.

\bibitem{swad}
J.~Cha, S.~Chun, K.~Lee, H.-C. Cho, S.~Park, Y.~Lee, and S.~Park, ``Swad:
  Domain generalization by seeking flat minima,'' in \emph{Advances in Neural
  Information Processing Systems (NeurIPS)}, 2021.

\bibitem{MAML}
C.~Finn, P.~Abbeel, and S.~Levine, ``Model-agnostic meta-learning for fast
  adaptation of deep networks,'' in \emph{ICML}, 2017, pp. 1126--1135.

\bibitem{TID2020}
G.~Wang, C.~Luo, X.~Sun, Z.~Xiong, and W.~Zeng, ``Tracking by instance
  detection: A meta-learning approach,'' in \emph{CVPR}, 2020, pp. 6287--6296.

\bibitem{BridgingDT2019}
L.~Huang, X.~Zhao, and K.~Huang, ``Bridging the gap between detection and
  tracking: A unified approach,'' in \emph{ICCV}, 2019, pp. 3998--4008.

\bibitem{generalReid}
Y.~Dai, X.~Li, J.~Liu, Z.~Tong, and L.-Y. Duan, ``Generalizable person
  re-identification with relevance-aware mixture of experts,'' in \emph{CVPR},
  2021, pp. 16\,145--16\,154.

\bibitem{Person30K}
Y.~Bai, J.~Jiao, C.~Wang, J.~Liu, Y.~Lou, X.~Feng, and L.~yu~Duan,
  ``{Person30K}: A dual-meta generalization network for person
  re-identification,'' in \emph{CVPR}, 2021, pp. 2123--2132.

\bibitem{3DVField}
A.~Lehner, S.~Gasperini, A.~Marcos-Ramiro, M.~Schmidt, M.-A.~N. Mahani,
  N.~Navab, B.~Busam, and F.~Tombari, ``3d-vfield: Learning to adversarially
  deform point clouds for robust 3d object detection,'' \emph{arXiv preprint
  arXiv:2112.04764}, 2021.

\bibitem{pointaugment}
R.~Li, X.~Li, P.-A. Heng, and C.-W. Fu, ``Pointaugment: an auto-augmentation
  framework for point cloud classification,'' in \emph{Proceedings of the
  IEEE/CVF Conference on Computer Vision and Pattern Recognition}, 2020, pp.
  6378--6387.

\end{thebibliography}
\end{document}